\documentclass{article}

\PassOptionsToPackage{numbers, compress}{natbib}

\usepackage[dblblindworkshop, final]{neurips_2026}
\workshoptitle{NeurIPS 2026 Workshop on On-Device Intelligence: Foundation Models under Real-World Constraints}

\makeatletter
\renewcommand{\@trackname}{\@workshoptitle.}
\makeatother

\usepackage[utf8]{inputenc} 
\usepackage[T1]{fontenc}    
\usepackage{hyperref}       
\usepackage{url}            
\usepackage{booktabs}       
\usepackage{amsfonts}       
\usepackage{nicefrac}       
\usepackage{microtype}      
\usepackage{xcolor}         
\hypersetup{hidelinks}

\usepackage{amsmath} 
\usepackage{amssymb}
\usepackage{optidef}
\usepackage{xfrac}
\usepackage{cancel}
\usepackage{graphicx}

\newcommand{\W}{\mathbf{W}}
\newcommand{\Sp}{\mathbf{S}}
\newcommand{\Q}{\mathbf{Q}}

\newcommand{\Lo}{\mathbf{L}}

\usepackage{cleveref}
\usepackage{lipsum}
\usepackage{comment}
\usepackage{pifont}
\usepackage{bbding}
\usepackage{svg}
\usepackage{multirow}
\usepackage{booktabs}
\usepackage{graphicx}
\usepackage{array}
\usepackage{makecell}
\usepackage[inline]{enumitem}
\usepackage{float}
\usepackage{algorithm}
\usepackage[indLines]{algpseudocodex}
\usepackage[table]{xcolor}

\newcommand{\ignore}[1]{}
\newcommand{\pb}[1]{\cellcolor{blue!10}}

\title{Quantization of Large Models \\ with Weight Outliers and Sparsity}

\author{%
\begin{tabular}{c}
Leonidas Tsigkounakis$^{\star}$ \qquad Abraham Jaeger Mountain$^{\star}$ \\
Yilang Zhang$^{\star\dagger}$ \qquad Georgios B. Giannakis$^{\star}$
\end{tabular}
\vspace{10pt}
\\
$^{\star}$Department of ECE, University of Minnesota, Minneapolis, MN 55455\\
$^{\dagger}$ML Research, Morgan Stanley, New York, NY 10019
}

\begin{document}

\maketitle

\begin{abstract}
Post-training quantization (PTQ) reduces the memory burden of large model deployment by casting the high precision trained weights to fewer bits. While this promotes memory-efficient inference, it also incurs quantization error that impairs model performance. To compensate for quantization error, the present contribution decomposes the original trained weights into a quantized matrix, a low-rank component, and a sparse matrix that accounts for outlier rows in the weights. Due to analytical intractability and the discrete structure of the decomposition, a simulated annealing search is employed to explore candidate decompositions while avoiding poor local minima. Numerical tests demonstrate the effectiveness of this approach across a broad suite of large language models, including Llama 3.2 1B, Llama 3.2 3B, Llama 3.1 8B, and Qwen2.5 7B, evaluated on natural language generation benchmarks using 4-bit NF4 and FP4 quantizers. The novel method can markedly outperform direct quantization while incurring minimal memory overhead.
\end{abstract}

\section{Introduction}
\label{sec:intro}

Large language models (LLMs)~\cite{jiangMistral7B2023, grattafioriLlama3Herd2024} have demonstrated remarkable effectiveness across a wide range of tasks, including natural language understanding~\cite{NEURIPS2019_4496bf24}, code generation~\cite{chenEvaluatingLargeLanguage2021}, and reasoning~\cite{biskPIQAReasoningPhysical2020}. Underpinning these capabilities are models with billions of parameters, typically represented in 16-bit floating-point formats (e.g. FP16, BF16). Although training such models requires substantial computational and energy resources, this process is typically performed once. In contrast, the subsequent LLM deployment requires sustained memory and bandwidth throughout the model's lifetime. As models scale, memory-efficient deployment has become an increasingly important challenge, particularly in resource-constrained and latency-sensitive environments such as edge devices and the cloud.

To address this, several model compression techniques have been advocated. For example; pruning~\cite{ashkboos2024slicegpt} reduces the parameter count by removing weights; knowledge distillation~\cite{fangKnowledgeDistillationDataset2025} transfers knowledge to a smaller student model; and quantization~\cite{NEURIPS2022_c3ba4962, frantarGPTQAccurateQuantization2023} maps model parameters to lower-precision numerical formats. Among these approaches, low-bit quantization is particularly attractive because it can markedly curtail memory footprint and bandwidth requirements. 

Post-training quantization (PTQ) is a popular paradigm because it applies quantization to a model after training, as opposed to alternative paradigms that incorporate quantization during pretraining~\cite{liu-etal-2024-llm} or fine-tuning~\cite{wei2025roste}. Conventional PTQ typically leverages a calibration dataset to determine the quantization parameters~\cite{frantarOptimalBrainCompression2022, frantarGPTQAccurateQuantization2023}, whereas zero-shot PTQ directly quantizes model parameters without additional data. This distinction establishes zero-shot PTQ as remarkably attractive in scenarios where representative calibration data is unavailable. However, a central challenge in PTQ is coping with the associated quantization error, especially when compressing to 4 bits or less. For LLMs, the performance impairment can be exacerbated because errors propagate in the transformer layers and across output tokens in long-sequence inference~\cite{NEURIPS2022_c3ba4962, dettmersSpQRSparsequantizedRepresentation2024}.

A major source of quantization error is the presence of high-magnitude outliers in LLM weight distributions, which can exhibit structured patterns~\cite{NEURIPS2022_c3ba4962, dettmersSpQRSparsequantizedRepresentation2024}\ignore{, which investigates various LLM outlier types}. Directly quantizing such outliers is challenging because they enlarge the quantization range and thereby reduce quantization fidelity for the majority of weights, especially in low-bit regimes. As a result, outliers can contribute disproportionately to quantization error and lead to downstream performance degradation. Prior work has proposed several strategies to mitigate this issue. For instance, Gpt3.int8()~\cite{NEURIPS2022_c3ba4962} and SPQR~\cite{dettmersSpQRSparsequantizedRepresentation2024} isolate high-magnitude activation or weight outliers and retain them in higher precision. SmoothQuant~\cite{xiao2023smoothquant} transfers activation outliers to weights using per-channel scaling, while AWQ~\cite{MLSYS2024_42a452cb} performs saliency-aware weight quantization. QuaRot~\cite{NEURIPS2024_b5b93943} instead applies rotation matrices to redistribute activation magnitude.  Many of these approaches, however, rely on calibration data and can thus be less suitable in fully calibration-free/zero-shot deployment settings.

Rather than directly quantizing pretrained weights, we adopt a calibration-free mixed-precision decomposition framework that factorizes high-precision pretrained weights into three complementary components: a quantized backbone, a high-precision sparse component, and a high-precision low-rank factor. This decomposition is motivated by the observation that, once the high-energy outliers are isolated, the remaining reconstruction error often admits an efficient low-rank approximation capturing the dominant singular directions. Together, the sparse and low-rank components improve reconstruction fidelity while incurring minimal memory overhead and inference latency. Since a global search for an optimal decomposition scales exponentially with weight matrix dimensions, the present contribution leverages a simulated annealing algorithm to explore candidate decompositions while avoiding poor local minima. Experiments on zero-shot natural language generation tasks demonstrate the effectiveness of the proposed approach across a wide range of popular large language models quantized to 4-bit NF4 and FP4 formats.
\section{Preliminaries}
\label{sec:prelim}
\begin{figure*}[t] 
    \centering
    \includegraphics[width=0.9\textwidth]{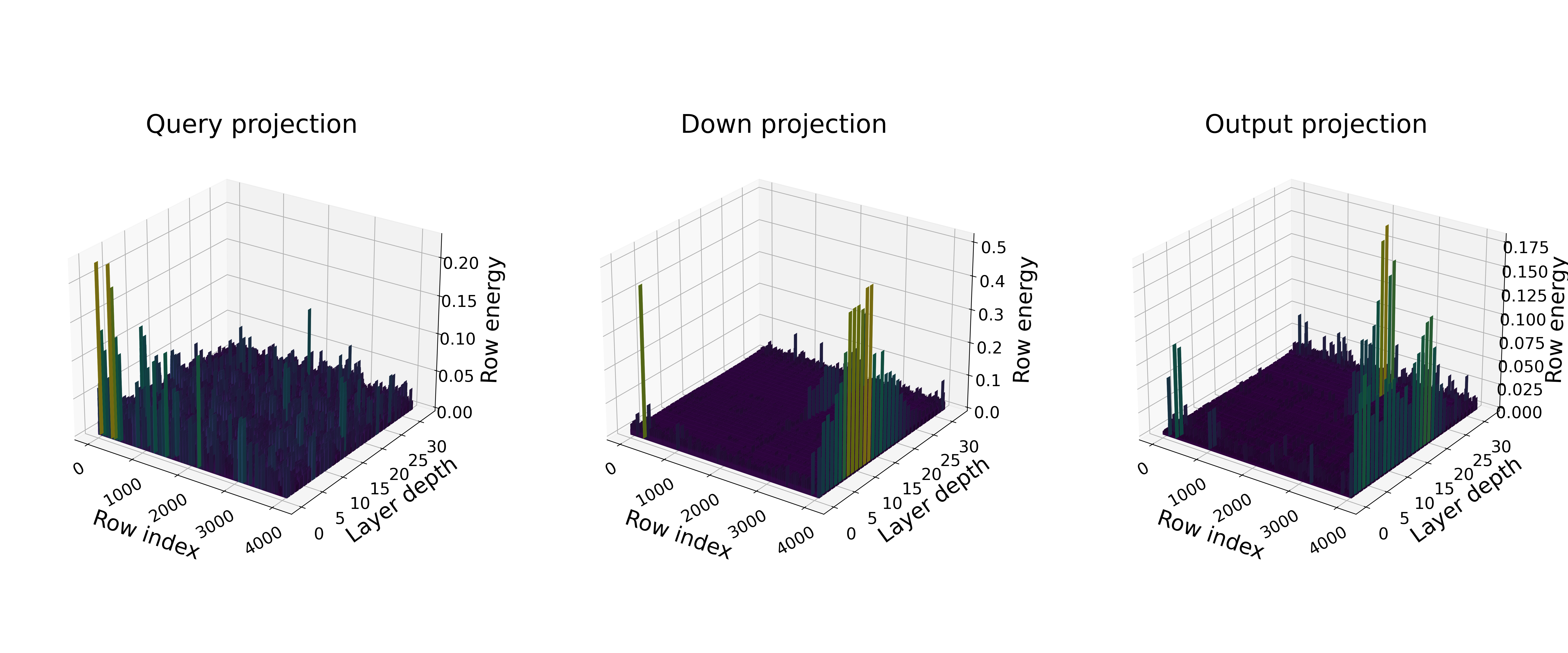}
    \vspace{-0.8cm}
    \caption{From left to right: squared $\ell_2$ row norms of the quantization error (NF4, block size 64) for the pretrained query, down, and output projection matrices of Llama 3.1 8B, plotted against layer depth. The quantization error exhibits structured high-energy outlier rows across several layers.}
    \label{fig:3d}
\end{figure*}
Consider a general weight matrix $\W\in\mathbb{R}^{m\times n}$ from a pretrained model stored in 16-bit floating-point precision. To reduce the memory footprint during deployment, a quantizer $q_b: \mathbb{R}^{m\times n} \to \mathcal{C}_{b}^{m\times n}$ maps entries of $\W$ onto a finite $b$-bit codebook $\mathcal{C}_{b} \subset \mathbb{R}$ and stores the resulting low-precision representation $\Q := q_b(\W)$  in memory. The stored values are converted back to a higher-precision compute format on the fly through a dequantizer $d_b: \mathcal{C}_{b}^{m\times n} \to \mathbb{R}^{m\times n}$. The overall quantize-then-dequantize operator is denoted by $\hat{q}_b \coloneq d_b\circ q_b$, which is not an identity mapping owing to quantization error. Accordingly, the quantization error is $\mathbf{E} := \W - d_{b}(\Q)$,
whose magnitude depends on both the quantizer design and the distribution of the weight entries.

A representative example is normal float (NF) quantization~\cite{dettmersQLoRAEfficientFinetuning2023} that constructs a codebook with a standard Gaussian prior for the weight entries. Letting $\Phi^{-1}$ denote the quantile function of the standard Gaussian distribution $\mathcal{N}(0, 1)$, the normalized $b$-bit NF codebook is defined by
\[
\mathcal{C}_{b}
=\left\{
\frac{z_k}{\max_{1 \leq j \leq 2^b}|z_j|}
\right\}_{k=1}^{2^{b}}, \quad 
z_k = \frac{1}{2}\Phi^{-1}\!\left(\frac{k}{2^{b}+1}\right)
+ \frac{1}{2}\Phi^{-1}\!\left(\frac{k+1}{2^{b}+1}\right).
\]
Given a scaling factor $s = \max_{i,j}|w_{ij}| > 0$, the reconstruction is obtained via a nearest-neighbor solution in the normalized domain followed by rescaling
\[
[\hat{q}_{b}(\W)]_{ij}
= s \cdot \arg\min_{z \in \mathcal{C}_{b}} \left| s^{-1} w_{ij} - z \right|.
\]
That is, the high-precision weights are first normalized to $[-1,1]$, quantized to the nearest codebook level, and then mapped back to the original range by scaling with the largest absolute entry of the original weights. In low-bit regimes such as 4-bit quantization, NF4 (i.e. the 4-bit instance of NF quantization) has become a popular choice as it can yield lower quantization error than integer or floating-point alternatives. In practice, the weight matrix can be partitioned into $B$ non-overlapping blocks of equal size with scaling factor $s_{B}$ assigned block-wise. While this fine-grained approach can account for variation in weight magnitude, quantization remains sensitive to large-magnitude outliers, which can degrade the quantization fidelity around the majority of small weights.

Pretrained LLM weights, specifically, are observed to exhibit both entry-wise and row-wise outliers, i.e., entire rows with significantly larger entries. This outlier pattern propagates to the quantization error as well, even if quantization is performed block-wise. Figure~\ref{fig:3d} displays the row-wise energy $\|\mathbf{E}_{i,:}\|_{2}^{2}$ of the quantization error for the Llama 3.1 8B pretrained weights against layer depth. Direct NF4 weight quantization was applied with a block size of 64 elements. The figure corroborates that outliers concentrate in a small subset of rows, particularly in the multi-layer perceptrons (MLP) of the transformer blocks. Orthogonally, the quantization error can be effectively captured by a low-rank approximation. Although $\mathbf{E}$ is typically high-rank, it is observed that the leading singular values can be significantly larger than the remaining spectrum~\cite{saha_caldera_2024}. Prior work~\cite{liLoftQLoRAfinetuningawareQuantization2024} has also explored low-rank structure in the quantization error, albeit in a LoRA fine-tuning setting. In contrast, this work investigates outlier-aware zero-shot PTQ, where calibration or fine-tuning data may be unavailable. This motivates a co-design of outlier modeling and low-rank approximation for quantization error.

\textbf{Notation. } Calligraphic uppercase letters denote sets. Bold uppercase letters denote matrices, while bold lowercase characters, vectors. For some matrix $\mathbf{Z}\in\mathbb{R}^{m\times n}$, $\mathbf{Z}_{i,:}$ denotes its $i$-th row, $\|\mathbf{Z}\|_{\mathrm{F}}$ denotes the Frobenius norm, and $\|\mathbf{Z}\|_{2,0}$ counts the number of nonzero rows. The symbol $\circ$ denotes function composition, and $\mathrm{SVD}_r(\mathbf{Z}) := \mathbf{U}_r\mathbf{\Sigma}_r\mathbf{V}_r^{\top}$ denotes the rank-$r$ truncated SVD approximation of $\mathbf{Z}$. The symbol $\mathbb{I}(\cdot)$ is the indicator function, while $H_\kappa:\mathbb{R}^{m\times n}\mapsto \mathbb{R}^{m\times n}$ denotes the hard thresholding function $H_{\kappa}(\mathbf{Z})$ that preserves the $\kappa \in \{ 0, \dots, m \}$ highest energy rows of $\mathbf{Z}$, while setting the rest to $\mathbf{0}_n^\top$.
\section{Outlier-aware zero-shot PTQ via simulated annealing}
\label{sec:method}

To cope with the quantization error under row outlier patterns, this paper advocates decomposing the trained weights into a quantized matrix $\Q\in\mathcal{C}_b^{m\times n}$, a row-sparse matrix $\Sp\in\mathbb{R}^{m\times n}$, and a low-rank matrix $\Lo \in\mathbb{R}^{m\times n}$ with rank $r \ll\min(m, n)$.
This mixed-precision decomposition of $\W$ can be captured via the discrete optimization problem 
\begin{align}
    \label{eq:rosann}\nonumber
    \min_{\Q \in \mathcal{C}_{b}^{m \times n}}\min_{\Sp, \Lo} \quad & \| \W - \Sp - d_{b}(\Q) - \Lo \|^2_{\mathrm{F}} 
    \\ \mathrm{s.t.} \quad 
    & \mathrm{rank}(\Lo) \leq r 
    \\ \quad 
    & \| \Sp \|_{2,0} \leq \kappa
    \nonumber
\end{align}
where $\|\mathbf S\|_{2,0} = \sum_{i=1}^m \mathbb I\{\|\mathbf S_{i,:}\|_2 > 0\}$, and $\kappa \ll m$. Although the cost function in~\eqref{eq:rosann} is analytically tractable, the caveat is that the search space induced by $\Q$ has cardinality $2^{bmn}$. This renders an exhaustive combinatorial search for $\Q$ computationally intractable with $\mathbf{S}$ and $\Lo$ given. An alternative is to adopt block coordinate descent solvers cycling between each component. However, such methods are prone to poor local minima in this highly non-convex and discrete search space, and lack a principled mechanism for global exploration.

To address this limitation, this paper adopts a meta-heuristic \textit{simulated annealing} algorithm, which performs a search over candidate decompositions using temperature-controlled random perturbations. The temperature follows a pre-defined schedule that decreases over time, thus encouraging exploration of the quantized matrix set $\mathcal{C}_b^{m\times n}$ to expose non-local but potentially superior $\mathbf{Q}$ candidates.

Specifically, the quantized backbone $\mathbf{Q}$ is initialized as the direct quantization \(\mathbf{Q}^{(0)} = q_b(\mathbf{W})\),
and the row-sparse component is initialized as $\mathbf{S}^{(0)}=\mathbf{0}_{m\times n}$. Given a decreasing temperature schedule $\{ \theta_k \}_{k=0}^{K-1}$ (e.g. linear, geometric, cosine), the simulated annealing search generates proposals
\begin{align*}
   &\boldsymbol{\Pi}_{k+1} \sim f(\boldsymbol{\Pi}_{k+1} ;\;\theta_{k} ) \\
   &\check{\mathbf{L}}^{(k+1)} = \mathrm{SVD}_r\bigl(\mathbf{W} - d_{b}(\mathbf{Q}^{(k)}) + \boldsymbol{\Pi}_{k+1} \bigr), \\
   &\check{\mathbf{Q}}^{(k+1)} = q_b \bigl( \mathbf{W} - \check{\mathbf{L}}^{(k+1)} - \mathbf{S}^{(k)} \bigr), \\
   &\check{\mathbf{S}}^{(k+1)} = H_{\kappa}\bigl(\mathbf{W} - d_{b}(\check{\mathbf{Q}}^{(k+1)}) - \check{\mathbf{L}}^{(k+1)} \bigr)
\end{align*}
for $k=0,\dots,K-1$. $\boldsymbol{\Pi}_{k+1} \in \mathbb{R}^{m\times n}$ is sampled from  a pre-specified prior $f(\boldsymbol{\Pi}_{k+1};\; \theta_{k})$, which can be a Gaussian distribution with standard deviation scaled by $\theta_{k}$, for instance. This perturbation in the low-rank component encourages exploration of $\mathbf{Q}$ in regions where the error is approximately low rank while avoiding being trapped around the current iterate $\mathbf{Q}^{(k)}$. The cost of the current proposal is $\check{c}_{k+1} = \| \mathbf{W} - \check{\mathbf{S}}^{(k+1)} - d_{b}(\check{\mathbf{Q}}^{(k+1)}) - \check{\mathbf{L}}^{(k+1)} \|_\mathrm{F}^2$,
and the next iterate is taken as
\begin{align*}
    &(\mathbf{L}^{(k+1)},\; \mathbf{Q}^{(k+1)},\; \mathbf{S}^{(k+1)}) = \\
    &~~~~~~~~~~~~~~~~~~~~~=\begin{cases}
    \begin{aligned}
    &(\check{\mathbf{L}}^{(k+1)},\; \check{\mathbf{Q}}^{(k+1)},\; \check{\mathbf{S}}^{(k+1)})\quad\text{w.p.}\; P_{\text{accept}}, \end{aligned}\\
    \begin{aligned}
    &(\mathbf{L}^{(k)},\; \mathbf{Q}^{(k)},\; \mathbf{S}^{(k)})\; \quad \text{w.p.}\; 1 - P_{\text{accept}}
    \end{aligned}
    \end{cases}
\end{align*}
where the acceptance probability is $P_{\mathrm{accept}} = \min \{ e^{-(\check{c}_{k+1} - c_k)/\theta_{k}} , \,1\}$.
Note that a better proposal is always accepted, but a nominally worse proposal can be accepted with probability $P_{\mathrm{accept}}$, enabling the trajectory to escape sub-optimal local minima. The algorithm outlining this procedure can be found in section~\ref{pseudocode} of the Appendix.

At the final iteration $K$, the algorithm outputs a quantized matrix $\Q_f := \mathbf{Q}^{(K)}$, together with high-precision row-sparse and low-rank components $\mathbf{S}_f := \mathbf{S}^{(K)}$ and $\Lo_f := \mathbf{L}^{(K)}$, corresponding to the final accepted iterate. To reduce the additional memory and compute cost of storing the high-precision components densely, the row sparsity of $\mathbf{S}_f$ is exploited and the low-rank component is factorized with truncated rank-$r$ SVD. In particular, writing $\text{SVD}_r(\Lo_f) = \mathbf{U}_r\mathbf{\Sigma}_{r}\mathbf{V}_r^{\top}$, the low-rank component is stored through the factors $\Lo_1 := \mathbf{U}_r\mathbf{\Sigma}_{r} \in \mathbb{R}^{m\times r}$ and $\Lo_2 := \mathbf{V}_r \in \mathbb{R}^{n\times r}$ where $\Lo_f = \Lo_1\Lo_2^{\top}$. The row-sparse component is stored only through its nonzero rows and their indices. Letting $\mathcal{K}\subseteq\{1,\dots,m\}$ denote the index set of kept rows, with $|\mathcal{K}|=\kappa$, this reduces the storage of $\mathbf{S}_f$ from $mn$ to $\kappa n$ high-precision entries along with the negligible $\kappa$ row indices. Therefore, for an input $\mathbf{x}\in\mathbb{R}^{n}$, the per-matrix inference becomes $\mathbf{x}\W^{\top} \approx \mathbf{x}\Q_f^{\top} + \mathbf{x}\mathbf{S}_{f}^{\top} + (\mathbf{x}\Lo_2)\Lo_1^{\top}$, where the term $\mathbf{x}\mathbf{S}_f^{\top}$ can be evaluated only on the rows indexed by $\mathcal{K}$.

\begin{table*}[t]
\centering
\caption{Performance comparison on NLG benchmarks using select Llama 3 models and Qwen 2.5 7B. Word perplexity is reported on WikiText-2 (WT-2), while GSM8K and MBPP report exact-match accuracy and pass@1 code-generation accuracy, respectively. Higher is better for GSM8K and MBPP, while lower is better for WT-2.}
\vspace{5pt}
\small
\setlength{\tabcolsep}{3pt}
\renewcommand{\arraystretch}{1.0}

\begin{tabular}{@{}llcccccc@{}}
\toprule
\multirow{2}{*}{\textbf{Model}} & \multirow{2}{*}{\textbf{Method}} & \multicolumn{3}{c}{\textbf{NF4}} & \multicolumn{3}{c}{\textbf{FP4}} \\ 
\cmidrule(lr){3-5} \cmidrule(lr){6-8}
& & \textbf{GSM8K ($\uparrow$)} & \textbf{MBPP ($\uparrow$)} & \textbf{WT-2 ($\downarrow$)} & \textbf{GSM8K ($\uparrow$)} & \textbf{MBPP ($\uparrow$)} & \textbf{WT-2 ($\downarrow$)} \\
\midrule
\multirow{5}{*}{Llama 3.1 8B}
&\pb{}Original (FP16) & \pb{}50.27 & \pb{}47.60 & \pb{}7.55 & \pb{}50.27 & \pb{}47.60 & \pb{}7.55 \\
& QLoRA init. & 44.73 & 47.00 & 8.04 & 38.29 & 43.40 & 8.51 \\
& LoftQ init.    & 44.12 & 44.80 & 7.98 & 38.67 & 44.20 & 8.39 \\
& Ours$_{\rho=0\%}$      & 44.96 & 46.80 & 7.93 & 42.68 & 45.60 & 8.32 \\
& Ours$_{\rho=1\%}$   & \textbf{46.32} & \textbf{48.00} & \textbf{7.89} & \textbf{43.97} & \textbf{46.40} & \textbf{8.31} \\
\midrule
\multirow{5}{*}{Qwen 2.5 7B}
& \pb{}Original (FP16) & \pb{}83.40 & \pb{}63.40 & \pb{}8.74 & \pb{}83.40 & \pb{}63.40 & \pb{}8.74 \\
&  QLoRA init.  & 78.70 & 61.40 & 9.09 & 79.76 & \textbf{62.40} & 9.61 \\
& LoftQ init.    & 79.91 & 62.60 & 9.06 & 79.38 & 61.80 & 9.47 \\
& Ours$_{\rho=0\%}$     & 81.50 & 60.00 & 9.01 & 77.03 & 57.40 & 9.31 \\
& Ours$_{\rho=1\%}$   & \textbf{82.49} & \textbf{62.80} & \textbf{9.01} & \textbf{80.14} & 62.20 & \textbf{9.30} \\
\midrule
\multirow{5}{*}{Llama 3.2 3B}
& \pb{}Original (FP16) & \pb{}27.45 & \pb{}38.60 & \pb{}9.51 & \pb{}27.45 & \pb{}38.60 & \pb{}9.51 \\
&  QLoRA init.  & 22.52 & 32.60 & 9.98 & 18.73 & 33.40 & 10.42 \\
& LoftQ init.    & 21.53 & 34.20 & 9.93 & 19.03 & 35.00 & 10.30 \\
& Ours$_{\rho=0\%}$      & \textbf{24.72} & \textbf{37.80} & \textbf{9.83} & 19.56 & \textbf{36.00} & 10.20 \\
& Ours$_{\rho=1\%}$   & 23.58 & 35.00 & 9.86 & \textbf{22.21} & 31.60 & \textbf{10.18} \\
\midrule
\multirow{5}{*}{Llama 3.2 1B}
& \pb{}Original (FP16) & \pb{}5.61 & \pb{}26.60 & \pb{}8.07 & \pb{}5.61 & \pb{}26.60 & \pb{}8.07 \\
& QLoRA init.  & 2.81 & 18.40 & 13.06 & 1.36 & 13.40 & 14.28 \\
& LoftQ init.    & 3.71 & 21.60 & 12.92 & 2.27 & 10.20 & 13.77 \\
& Ours$_{\rho=0\%}$      & 3.94 & 19.40 & 12.73 & 2.20 & \textbf{15.40} & 13.49 \\
& Ours$_{\rho=1\%}$   & \textbf{4.70} & \textbf{22.20} & \textbf{12.69} & \textbf{2.65} & 15.20 & \textbf{13.42} \\
\bottomrule
\end{tabular}
\label{tab:tab}
\end{table*}

\section{Numerical Tests}
\label{sec:tests}

This section assesses the proposed zero-shot PTQ method on Llama 3.2 1B, Llama 3.2 3B, Llama 3.1 8B, and Qwen 2.5 7B models, which are evaluated with the LM Evaluation Harness API~\cite{eval-harness} on GSM8K~\cite{cobbeTrainingVerifiersSolve2021}, MBPP~\cite{austinProgramSynthesisLarge2021}, and WikiText-2~\cite{merity2017pointer} datasets. Direct quantization, LoFTQ initialization~\cite{liLoftQLoRAfinetuningawareQuantization2024}, and the proposed method are compared under both NF4~\cite{dettmersQLoRAEfficientFinetuning2023} and FP4 quantization. All experiments use a rank $r=32$ for $\mathbf{L}$, and $K=300$ annealing iterations. To control the row sparsity, a percentage-wise row budget $\rho$ is defined a-priori, where $\kappa = \lfloor \rho m\rfloor$. Both $\rho =0\%$ (i.e. $\Sp=\mathbf{0}_{m\times n}$) and $\rho=1\%$ are evaluated. The decomposition is applied to all transformer attention and MLP projections (i.e. \texttt{q\_proj}, \texttt{k\_proj}, \texttt{v\_proj}, \texttt{o\_proj}, \texttt{up\_proj}, \texttt{down\_proj}, and \texttt{gate\_proj}). Projections of the same type share a common geometrically decreasing annealing temperature schedule, obtained from a Bayesian optimization pre-processing step (cf.~\cite{garnett_bayesoptbook_2023}). A single NVIDIA A100 40GB GPU is employed for all tests.

The results are summarized in Table~\ref{tab:tab}, comparing direct zero-shot quantization, the LoFTQ initialization, annealing without row sparsity ($\rho =0\%$), and the proposed row-sparse annealing method with $\rho =1\%$. Although LoFTQ~\cite{liLoftQLoRAfinetuningawareQuantization2024} was originally proposed for LoRA-aware quantization, we include it as a baseline under the same zero-shot PTQ evaluation protocol. Across NF4 and FP4 quantization, the proposed row-sparse plus annealing method achieves the best or near-best performance in most settings, and the gains are especially consistent on WikiText-2, where it obtains the lowest perplexity for every model and quantization format except for the NF4 Llama 3.2 3B case. This suggests that explicitly modeling row-structured outliers can better preserve language-modeling performance under low-bit quantization. In GSM8K, it improves over direct quantization in all cases and typically outperforms both the LoFTQ initialization and annealing without row sparsity, indicating that the sparse correction is beneficial for preserving mathematical reasoning performance under low-bit quantization. 

For code generation with MBPP, our methods achieve the best pass@1 score for NF4 quantization across all four models. With FP4, however, plain quantization is slightly better on Qwen 2.5 7B, and annealing without row sparsity performs better on Llama 3.2 3B and Llama 3.2 1B. This could suggest that the benefits of the low-rank and row-sparse corrections can overlap, and the design choice of explicit outlier modeling depends on the model, task, and structure of the quantization error. Overall, the results support that annealing-based decompositions improve over alternative approaches, and adding row-sparse outlier absorption can boost gains with minimal storage overhead (cf. Appendix~\ref{memory}), particularly for perplexity and GSM8K accuracy.
\section{Conclusion and future work}
\label{sec:conclusion}
This paper introduced a calibration-free post-training quantization method that decomposes pretrained LLM weights into quantized, low-rank, and row-sparse components. By explicitly modeling row-structured outliers in the weights, the proposed decomposition improves zero-shot performance over direct quantization and low-rank alternatives across several models and quantization formats. These results suggest that mixed-precision decompositions can be beneficial for low-bit PTQ, particularly when calibration data or fine-tuning data is unavailable. Future work will investigate using the low-rank factors as an initialization for LoRA fine-tuning, exploring whether zero-shot performance correlates with downstream adaptation capacity.

\clearpage
\bibliographystyle{IEEEbib}
\bibliography{strings,refs}

\newpage
\appendix
\section{Algorithm pseudocode} \label{pseudocode}
This Appendix section provides the pseudocode of the present contribution in Algorithm~\ref{alg:annealing-rs}. The algorithm inputs are the pretrained high-precision weights $\mathbf{W}$, the number of annealing steps $K$, the rank $r$ of the low-rank component $\mathbf{L}$, the sparsity-controlling parameter $\kappa$ for the row-sparse $\mathbf{S}$, and the temperature schedule $\theta_{k} := \theta_{0} (\theta_{K-1}/{\theta_{0}})^{{k}/{(K-1)}}$. While the temperatures are geometrically decreasing, alternative schedules such as cosine or linear can be readily employed. Another design choice concerns the perturbation distribution, which in this work is defined as a zero-mean Gaussian with standard deviation proportional to the temperature. We leave alternative schedule and perturbation design choices for future research. 

\begin{algorithm}[h]
    \caption{Simulated annealing solver for quantized, low-rank, and row-sparse decomposition}\label{alg:annealing-rs}
    \begin{algorithmic}[1]
        \State \textbf{Input:} $\mathbf{W}, K, \theta_0, \theta_{K-1}, r, \kappa$
        
        \State \textbf{Initialize:} 
        $\mathbf{Q}^{(0)} = q_b(\mathbf{W})$,
        $\mathbf{S}^{(0)} = \mathbf{0}_{m\times n}$,
        $c_0 = +\infty$
        \For{$k = 0, \dots, K - 1$}
            \State $\theta_{k} = \theta_{0}(\theta_{K-1}/\theta_{0})^{k/(K-1)}$
            
            \State sample $[\boldsymbol{\Pi}_{k+1}]_{ij} \sim \mathcal{N}\!\left(0,\theta_{k}^{2}/4\right),\ \forall i,j$
            
            \State $\check{\mathbf{L}}^{(k+1)} = \mathrm{SVD}_r\bigl(\mathbf{W} - d_{b}(\mathbf{Q}^{(k)}) + \boldsymbol{\Pi}_{k+1} \bigr)$
            
            \State $\check{\mathbf{Q}}^{(k+1)} = q_b \bigl( \mathbf{W} - \check{\mathbf{L}}^{(k+1)} - \mathbf{S}^{(k)} \bigr)$
            
            \State $\check{\mathbf{S}}^{(k+1)} = H_{\kappa}\bigl(\mathbf{W} - d_{b}(\check{\mathbf{Q}}^{(k+1)}) - \check{\mathbf{L}}^{(k+1)} \bigr)$
            
            \State $\check{c}_{k+1} = \| \mathbf{W} - \check{\mathbf{S}}^{(k+1)} - d_{b}(\check{\mathbf{Q}}^{(k+1)}) - \check{\mathbf{L}}^{(k+1)} \|_\mathrm{F}^2$

            \State $P_{\mathrm{accept}} = \min \{ e^{-(\check{c}_{k+1} - c_k)/\theta_{k}} , \,1\}$
            
            \State sample $u \sim \text{Uniform}[0,1]$
            
            \If{$P_{\mathrm{accept}} > u$}
            
                \State \((\mathbf{L}^{(k+1)},\; \mathbf{Q}^{(k+1)},\; \mathbf{S}^{(k+1)}) = (\check{\mathbf{L}}^{(k+1)},\; \check{\mathbf{Q}}^{(k+1)},\; \check{\mathbf{S}}^{(k+1)})\)
                
                \State $c_{k+1} = \check{c}_{k+1}$
                
            \Else
            
                \State \((\mathbf{L}^{(k+1)},\; \mathbf{Q}^{(k+1)},\; \mathbf{S}^{(k+1)}) = (\mathbf{L}^{(k)},\; \mathbf{Q}^{(k)},\; \mathbf{S}^{(k)})\)
                
                \State $c_{k+1} = c_k$
            \EndIf
        \EndFor
        \State \textbf{Output:} $\Q^{(K)}, \Sp^{(K)}, \mathbf{L}^{(K)}$
    \end{algorithmic}
\end{algorithm}

\section{Memory overhead analysis} 
\label{memory}

The memory overhead is measured in bits per parameter, and is calculated for all seven attention and MLP projection types in every layer. The low-rank components $L_1$ and $L_2$ are stored in FP32 and employ a rank \(r=32\). The non-zero rows of each row-sparse matrix $S$ and their indices are stored in FP16 and FP32 respectively. For both the NF4 and FP4 quantization schemes, one FP32 scaling factor is used per block of 64 weights without double quantization. The remaining parameters are stored in FP16. Table~\ref{tab:storage} provides a summary of the results. 

\begin{table*}[ht]
\centering
\caption{Memory storage comparison across quantization schemes in bits per parameter. Values apply to both NF4 and FP4 quantization schemes.}
\vspace{5pt}
\small
\setlength{\tabcolsep}{3pt}
\renewcommand{\arraystretch}{1.0}
\begin{tabular}{@{}lccccc@{}}
\toprule
\textbf{Model} & \shortstack{\textbf{Original}\\\textbf{(FP16)}}
& \textbf{QLoRA init.} & \textbf{LoftQ init.}
& \textbf{Ours}$_{\rho=0\%}$ & \textbf{Ours}$_{\rho=1\%}$ \\
\midrule
Llama 3.1 8B & \pb{}16.000 & 6.005 & 6.339 & 6.339 & 6.477 \\
Qwen 2.5 7B & \pb{}16.000 & 6.146 & 6.486 & 6.486 & 6.621 \\
Llama 3.2 3B & \pb{}16.000 & 5.911 & 6.395 & 6.395 & 6.533 \\
Llama 3.2 1B & \pb{}16.000 & 6.945 & 7.529 & 7.529 & 7.653 \\
\bottomrule
\end{tabular}
\label{tab:storage}
\end{table*}
At a rank of \(r=32\), our method with \(\rho=1\%\) incurs an additional 0.12–0.14 bits per parameter relative to the \(\rho = 0\%\) or the alternating optimization cases. These estimates include quantization scales and row indices.

\end{document}